\documentclass[journal]{IEEEtran}

\usepackage{graphicx}
\usepackage{amsmath,amssymb,amsfonts}
\usepackage[hidelinks]{hyperref}
\usepackage{enumitem}
\usepackage{xcolor}
\usepackage{booktabs}
\usepackage{url}
\usepackage{tabularx}
\usepackage{array}
\usepackage[ruled,vlined,linesnumbered]{algorithm2e}
\usepackage{tikz}

\usetikzlibrary{arrows.meta,positioning,shapes.geometric,calc}
\usepackage{svg}
\usepackage{scalerel}
\usepackage{booktabs}
\usetikzlibrary{svg.path}
\definecolor{orcidlogocol}{HTML}{A6CE39}
\tikzset{
  orcidlogo/.pic={
    \fill[orcidlogocol] svg{M256,128c0,70.7-57.3,128-128,128C57.3,256,0,198.7,0,128C0,57.3,57.3,0,128,0C198.7,0,256,57.3,256,128z};
    \fill[white] svg{M86.3,186.2H70.9V79.1h15.4v48.4V186.2z}
                 svg{M108.9,79.1h41.6c39.6,0,57,28.3,57,53.6c0,27.5-21.5,53.6-56.8,53.6h-41.8V79.1z M124.3,172.4h24.5c34.9,0,42.9-26.5,42.9-39.7c0-21.5-13.7-39.7-43.7-39.7h-23.7V172.4z}
                 svg{M88.7,56.8c0,5.5-4.5,10.1-10.1,10.1c-5.6,0-10.1-4.6-10.1-10.1c0-5.6,4.5-10.1,10.1-10.1C84.2,46.7,88.7,51.3,88.7,56.8z};
  }
}
\newcommand\orcidicon[1]{\href{https://orcid.org/#1}{\mbox{\scalerel*{
\begin{tikzpicture}[yscale=-1,transform shape]
\pic{orcidlogo};
\end{tikzpicture}
}{|}}}}

\graphicspath{{./}{fig/}}

\begin{document}

\title{Large Language Model Assisted Operational Monitoring for Battery Energy Storage System Integrated Power Distribution Networks}

\author{Azmeer~Akhtar,
Md~Fazley~Rafy$^{\textsuperscript{\orcidicon{0000-0003-3057-9546}}}$\,,~\IEEEmembership{Member,~IEEE},
and~Anurag~K.~Srivastava$^{\textsuperscript{\orcidicon{0000-0003-3518-8018}}}$\,,~\IEEEmembership{Fellow,~IEEE}
\thanks{The authors are with the Lane Department of Computer Science and Electrical Engineering,
West Virginia University, Morgantown, WV 26505, USA.
This work was supported in part by the U.S. Department of Energy (DoE) and the Appalachian Regional Commission (ARC).}}

\maketitle

\IEEEpubid{\makebox[\columnwidth]{979-8-3315-5720-1/26/\$31.00 ©2026 IEEE\hfill}\hspace{\columnsep}\makebox[\columnwidth]{}} 
\IEEEpubidadjcol
\begin{abstract}
Battery energy storage systems (BESS) are increasingly used in distribution networks for voltage regulation and demand response, which increases the volume and complexity of operational telemetry available to grid operators. This paper presents an AI-enabled monitoring framework that connects a large language model (LLM) interface with a structured telemetry database for BESS-integrated distribution system analysis. Operator questions are submitted in natural language and translated into validated SQL queries using predefined database schema information and approved KPI views. Retrieved measurements, including bus voltages, state of charge, active power, and reactive power, are evaluated against engineering constraints for voltage limits, BESS operation, and demand response tracking. The framework is validated using hardware-in-the-loop co-simulation data from a BESS-equipped distribution feeder operating under reactive power-based voltage control and price-driven demand response. Case studies show that the framework generates valid database queries, identifies repeated voltage violations, detects reactive power overshoot, and evaluates active-power tracking performance. The results show that LLM-assisted monitoring can connect structured grid telemetry with automated engineering assessment for BESS operation analysis.
\end{abstract}


\section{Introduction}
\label{sec:intro}
\IEEEPARstart{D}{istribution} systems are undergoing a structural change as distributed energy resources (DERs), battery energy storage systems (BESS) deployed as non-wire alternatives (NWAs), inverter-based resources, and responsive loads become active participants in feeder operation \cite{fose2024empowering, 10318643}. Conventional distribution monitoring was designed primarily for radial power delivery, limited field measurements, and operator review of alarms from supervisory control and data acquisition systems. DER-rich feeders now exhibit bidirectional power flow, time-varying voltage profiles, rapid changes in net load, and control interactions among inverter-based assets \cite{sarwar2022characterization}. These operating conditions increase the amount of telemetry that grid operators must interpret when assessing voltage regulation, demand response performance, battery dispatch, and device-level constraint violations \cite{fose2024empowering}. BESS operation adds another layer of monitoring complexity because each storage asset couples electrical, control, and energy-state variables \cite{8013143}. Active power dispatch affects demand response and peak reduction objectives, while reactive power support, often managed through mechanisms such as droop control, affects voltage regulation and inverter capability limits \cite{wei2023chance}. The same BESS unit must also satisfy state-of-charge bounds, apparent-power capability constraints, ramp-rate limits, and control-mode requirements \cite{8013143}. As a result, operator assessment requires simultaneous interpretation of measured active power, measured reactive power, power references, bus voltages, state of charge, operating mode, and event timing. Manual inspection of these variables becomes inefficient when measurements are collected across multiple nodes, phases, devices, and control intervals. The growth of cyber-physical telemetry and IoT-based integration further increases the operator burden \cite{10918749}. Modern BESS-integrated testbeds and field deployments generate high-frequency data streams from phasor measurements, inverter controllers, battery management systems, communication networks, and event logs. Furthermore, as these systems become more reliant on communication networks, operators must monitor for cyber-resilience, distinguishing between normal physical grid fluctuations and cyber anomalies such as false data injection or communication delays \cite{10918749}. These data are often stored in structured databases, but operators typically need database-specific knowledge to retrieve the correct time windows, node identifiers, constraint flags, and aggregated performance metrics. This creates a gap between the availability of operational data and its practical use for real-time or post-event engineering assessment.
Large language models (LLMs) offer a potential interface layer between operator intent and structured grid telemetry, though their application requires strict guardrails to prevent unverified text generation \cite{11168242}. Instead of requiring an operator to manually write database queries, an LLM-assisted tool can translate natural-language questions into structured queries, retrieve relevant measurements, and summarize the results \cite{jin2024chatgrid}. For BESS-integrated distribution systems, this capability can support questions such as identifying nodes with repeated voltage violations, detecting reactive power overshoot during voltage-control operation, and evaluating active-power tracking during demand response events. To mitigate the risk of hallucination common in broad language models, the value of such a tool depends on controlled query generation, access to validated database views, and assessment logic grounded in strict power-system operating limits \cite{11168242, ruan2024applying}.

Traditional monitoring platforms address only part of this problem. Most existing tools rely on predefined dashboards, alarm lists, and manually configured database queries \cite{cobilean2023review}. These tools work well when the operator already knows which device, variable, and time window must be checked. They become less effective when BESS operation must be diagnosed across several related signals. A single operating event may involve bus voltage, active-power command, reactive-power command, measured inverter response, SOC, control mode, and event timing. Manual review of these variables can slow engineering assessment, especially when the data are distributed across ADMS, DERMS, historian, and database-centered monitoring systems \cite{spudic2022integration, wu2022distribution}. Prior work has improved DER and distribution-system observability through state estimation, event detection, anomaly detection, forecasting, and control-performance assessment \cite{sugunaraj2025distributed}. These methods automate specific tasks, but they usually operate as fixed-function modules \cite{kunkolienkar2024visualizing}. They do not convert an operator's natural-language question into a database operation. In this context, SQL-based analysis can provide a practical approach for connecting natural-language operator questions to structured grid telemetry to retrieve time windows, aggregate measurements, rank violations, and compute tracking errors from high-rate telemetry \cite{sugunaraj2025distributed}. However, SQL use requires knowledge of table schemas, joins, filters, aggregation functions, and variable definitions. This requirement creates a practical barrier between stored BESS telemetry and operator-facing engineering questions. Recent LLM applications in power systems have examined planning support, document retrieval, grid-code interpretation, contingency analysis, operator training, and natural-language interfaces for simulation and analytics workflows \cite{11168242, pnnl2024connecting, bonadia2023potential, jena2025llm}. These studies show that LLMs can help users access technical information and interact with computational tools. However, unrestricted LLM outputs can produce invalid commands or unsupported engineering statements when they are not tied to approved data sources and operating constraints \cite{11168242}. For BESS-integrated distribution monitoring, an unresolved gap remains. Existing tools do not provide a controlled natural-language interface that converts operator questions into validated read-only SQL, retrieves time-indexed cyber-physical telemetry, and evaluates the results against voltage, active-power, reactive-power, SOC, and demand-response constraints. This gap motivates the proposed LLM-assisted monitoring framework, which combines schema-constrained query generation, SQL validation, KPI-view execution, and engineering constraint assessment for BESS operation analysis.

\begin{figure*}[t]
\centering
\includegraphics[width=0.95\textwidth]{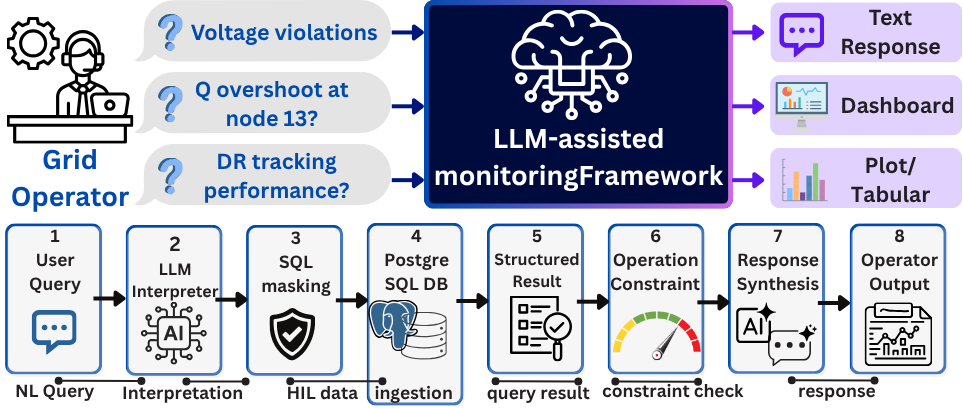}
\caption{LLM-assisted monitoring framework for BESS-integrated distribution system analysis. Operator questions are translated into validated SQL queries, evaluated against engineering constraints, and returned as text, dashboard, or plot/table outputs using operating data from the prior HIL testbed.}
\label{fig:architecture}
\end{figure*}

Without the LLM interface, these monitoring tasks require the operator to manually identify the relevant KPI view, select the correct node and time window, write SQL filters or joins, aggregate timestamped records, compare the returned values with operating limits, and translate the numerical output into an engineering assessment. This process is feasible for a database expert, but it slows operator-level analysis when questions combine voltage violations, reactive-power overshoot, active-power tracking, SOC limits, and data-confidence indicators. The proposed framework uses the LLM only as a controlled interface between operator intent and structured database queries; numerical evaluation remains governed by validated SQL execution and deterministic constraint checks. This paper addresses this need by presenting an AI-enabled monitoring framework that connects a natural-language operator interface to a structured PostgreSQL telemetry database for BESS-integrated distribution feeder analysis. The main contributions of this paper are as follows:
\begin{itemize}[leftmargin=*]
\item Developed a schema-constrained LLM-to-SQL monitoring pipeline for BESS-integrated distribution system telemetry. The pipeline converts natural-language operator questions into validated read-only SQL queries and restricts execution to predefined KPI views.
\item Integrated an engineering constraint assessment layer with the query pipeline to evaluate retrieved measurements against voltage regulation, active-power tracking, reactive-power tracking, SOC, and demand-response operating limits.
\item Validated using HIL co-simulation from a BESS-integrated distribution feeder under voltage-control and demand-response scenarios. The case studies evaluate voltage violation identification, reactive power overshoot detection, and active-power reference tracking from structured operational data.
\end{itemize}


\section{System Architecture and Methodology}

The proposed framework connects operator-facing natural-language queries with structured BESS telemetry stored in a PostgreSQL database. The architecture separates three functions: natural-language translation, database execution, and engineering constraint evaluation. This separation prevents the language model from directly interpreting raw telemetry without database validation or power-system constraint checks. The LLM does not access arbitrary database objects. Instead, it receives predefined schema context, summarized in Table~\ref{tab:feature_definitions}, and generates read-only SQL queries for approved KPI views. The generated query is validated before execution and thus grounds the natural-language interface in measured telemetry, predefined schema definitions, and operating constraints.

\subsection{Operational Constraint Framework}\label{sec:operational_flow}

The framework converts retrieved telemetry into engineering assessments using the deterministic constraint logic in Table~\ref{tab:constraints}. Retrieved measurements are evaluated against predefined power-system operating bounds rather than subjective language-model judgment \cite{walker2023battery}. Voltage regulation is assessed by comparing bus voltage with the allowable per-unit operating range. Active-power tracking is assessed by comparing measured BESS active power, $P_{\mathrm{meas}}$, with the active-power reference, $P_{\mathrm{ref}}$, during demand-response operation. Reactive-power response is assessed by comparing measured reactive power, $Q_{\mathrm{meas}}$, with the reactive-power reference, $Q_{\mathrm{ref}}$, and by checking reactive-power overshoot during voltage-control operation. Battery energy state is assessed using SOC limits to determine whether the BESS remains within its configured operating range. The framework also includes telemetry-confidence checks using the previously defined resiliency metric \cite{10918749}, when the corresponding fields are available. As shown in Table~\ref{tab:feature_definitions}, latency, anomaly flags, attack labels, and resilience scores can qualify the interpretation of retrieved measurements. These indicators help separate physical operating violations from intervals affected by delayed, stale, or abnormal telemetry.

\subsection{LLM-Assisted Query Pipeline}


In this implementation, the LLM module used \texttt{gpt-4o-mini} (OpenAI, 2024) with deterministic decoding settings to reduce output variability during SQL generation. Due to free tier account limitations, access was restricted to \texttt{gpt-4o-mini} (or legacy \texttt{gpt-3.5-turbo}) rather than flagship models such as GPT-5 or advanced reasoning tools. Algorithm~\ref{alg:llm_grid_query} summarizes the query generation and validation workflow. 
The prompt provided to the LLM included four controlled inputs: the operator’s natural-language question, the approved KPI view names \cite{hong2025next}, the schema fields listed in Table II, and explicit safety instructions requiring the output to be a single read-only PostgreSQL SELECT query. The model was instructed not to generate INSERT, UPDATE, DELETE, DROP, ALTER, CREATE, or multi-statement SQL commands, and not to reference database tables, views, columns, or joins outside the approved monitoring schema. The core prompt template was: ``Given the following approved BESS telemetry views and schema fields, generate one PostgreSQL SELECT query that answers the operator question. Use only the provided views and columns. Return SQL only. Do not modify the database.'' The generated SQL statement was then inspected by a rule-based validation layer before database execution. The validator rejected data-modification commands, unsupported views, unauthorized columns, multi-statement outputs, unsupported joins, and query patterns outside the approved monitoring scope. After validation, the SQL statement was executed against the PostgreSQL database, and the retrieved result set was evaluated by the engineering constraint layer described in Section \ref{sec:operational_flow}. The final output was formatted as a concise operational assessment for the operator.

\begin{table}[t]
\centering
\caption{Operational Constraint Framework for BESS Monitoring}
\label{tab:constraints}
\footnotesize
\begin{tabular}{p{0.20\columnwidth} p{0.35\columnwidth} p{0.27\columnwidth}}
\toprule
\textbf{Operational Metric} & \textbf{Evaluation Criteria} & \textbf{Assessment Output} \\
\midrule
Voltage Regulation & $V_{\min} \leq V_{bus} \leq V_{\max}$ \newline (e.g., 0.95 to 1.05 p.u.) & Violation count, vulnerable node ranking \\
\addlinespace
Active-Power Tracking & $|P_{meas} - P_{ref}| \leq \epsilon_{P}$ & Tracking error, DR performance status \\
\addlinespace
Reactive-Power Response& $|Q_{meas} - Q_{ref}| \leq \epsilon_{Q}$ \newline Check Q-overshoot & Reactive tracking status, overshoot flag \\
\addlinespace
Battery Energy State & $SOC_{\min} \leq SOC \leq SOC_{\max}$ & Available energy headroom, limit compliance \\
\addlinespace
Telemetry Confidence & Latency $\leq \tau_{\max}$ \newline Cyber-Anomaly Flag $= 0$ & Data-confidence qualifier \\
\bottomrule
\end{tabular}
\end{table}

\begin{table}[t]
\centering
\caption{PostgreSQL Schema Features for LLM Telemetry Queries}
\label{tab:feature_definitions}
\footnotesize
\begin{tabular}{@{} l l p{3.1cm} @{}}
\toprule
\textbf{SQL Column Name} & \textbf{Data Type} & \textbf{Description} \\
\midrule
\multicolumn{3}{@{}l}{\textit{Metadata and Spatiotemporal Identifiers}} \\
\texttt{ts}                 & Timestamp      & Time index for windowed SQL queries. \\
\texttt{run\_id}            & String         & Experimental scenario identifier. \\
\texttt{node\_id}           & String         & Feeder node for localized tracking. \\
\texttt{device\_id}         & String         & BESS or inverter endpoint identifier. \\
\texttt{control\_mode}      & Categorical    & E.g., Volt-VAR, Demand Response. \\
\midrule
\multicolumn{3}{@{}l}{\textit{Cyber-Physical Telemetry and States}} \\
\texttt{v\_bus\_pu}         & Numeric        & Per-unit bus voltage magnitude. \\
\texttt{p\_ref\_kw}, \texttt{p\_meas\_kw} & Numeric & Active-power command and response. \\
\texttt{q\_ref\_kvar}, \texttt{q\_meas\_kvar} & Numeric & Reactive-power command and response. \\
\texttt{soc\_percent}       & Numeric        & Battery state of charge available. \\
\texttt{price\_signal}      & Numeric        & Economic or DR dispatch trigger. \\
\midrule
\multicolumn{3}{@{}l}{\textit{Derived KPIs and Constraint Flags}} \\
\texttt{voltage\_viol\_flag}& Boolean        & Indicates $V_{bus}$ operating out of bounds. \\
\texttt{p\_tracking\_error} & Numeric        & Deviation between $P_{meas}$ and $P_{ref}$. \\
\texttt{max\_q\_overshoot}  & Numeric        & Peak reactive overshoot indicator. \\
\midrule
\multicolumn{3}{@{}l}{\textit{Cyber-Resilience and Trust Indicators}} \\
\texttt{latency\_ms}        & Numeric        & End-to-end communication delay. \\
\texttt{anomaly\_flag}      & Boolean        & Suspected cyber/physical anomaly. \\
\texttt{attack\_type}       & Categorical    & Delay, spoofing, or data manipulation. \\
\texttt{resilience\_score}  & Numeric        & Composite data confidence metric. \\
\bottomrule
\end{tabular}
\end{table}

\section{Cyber-Physical Co-Simulation Setup} 

The proposed framework is evaluated using time-series operating data generated from a real-time cyber-physical co-simulation environment. Rather than relying on static records, this setup couples feeder-level power-flow analysis with edge-controller communication and HIL-based inverter emulation to produce synchronized BESS operating measurements. The co-simulation architecture builds upon previously established frameworks for cyber-resilient BESS operation and edge-controller-based HIL validation \cite{10918749, 11457738}. The physical distribution network is modeled in OpenDSS using a modified IEEE 123-node unbalanced test feeder with a 4.16 kV primary system, where A BESS unit is connected at each of these secondary terminal buses. These units operate under voltage-control and demand-response scenarios, producing measured bus voltage, active-power response, reactive-power response, SOC, reference commands, and event-status records needed to evaluate the LLM-assisted monitoring framework. OpenDSS solves the feeder power flow and updates the network voltage response under changing BESS injections, while Typhoon HIL represents the localized BESS inverter response. 
The HIL model returns measured inverter outputs corresponding to the controller references. The communication layer uses Raspberry Pi single-board computers as edge controllers, replacing the earlier software-based network emulation approach. Each Raspberry Pi functions as an MQTT client and exchanges voltage measurements, active- and reactive-power setpoints, measured BESS responses, synchronization signals, and device-level records through assigned MQTT topics. The use of physical edge controllers allows the operating data to include communication delay, jitter, anomaly indicators, and data-confidence fields when available. During each simulation run, the MQTT message stream is captured and written to the PostgreSQL database using the structured schema in Table~\ref{tab:feature_definitions}. The database records preserve the timestamp, run identifier, simulation step, node identifier, BESS device identifier, feeder-bus and secondary-bus labels, control mode, voltage, active power, reactive power, SOC, price signal, demand-response flag, and derived KPI fields. The database also stores computed quantities such as voltage-violation flags, active-power tracking error, reactive-power tracking error, and maximum reactive-power overshoot. This data-recording process provides a clean, time-indexed operational database for the LLM-assisted query pipeline, while the LLM remains outside the closed-loop control process and is used only for post-simulation monitoring, SQL-based retrieval, and constraint-based engineering assessment.

\begin{algorithm}[t]
\DontPrintSemicolon
\SetAlgoLined
\SetAlgoNlRelativeSize{0}
\SetNlSty{textbf}{}{:}
\SetKwComment{Comment}{$\triangleright$~}{}
\SetKwInput{KwIn}{Input}
\SetKwInput{KwOut}{Output}
\caption{LLM-Driven Grid Operations Data Retrieval and Assessment}
\label{alg:llm_grid_query}
\footnotesize
\KwIn{Grid operator question $Q$; PostgreSQL database $\mathcal{D}$}
\KwOut{Textual assessment $T$; dashboard visualization $V$}
\BlankLine
Receive natural language question $Q$ from the frontend interface\;
\BlankLine
\Comment{LLM Agent Processing}
Analyze $Q$ to interpret the grid and battery query context\;
Generate SQL command $S$ from the interpreted context of $Q$\;
\BlankLine
\Comment{Database Execution}
Validate $S$ against approved views and permitted command types\;
Execute $S$ against $\mathcal{D}$; retrieve result set $R$\;
\BlankLine
\Comment{Constraint Evaluation and Assessment}
Evaluate $R$ against engineering constraint library\;
Format $R$ into time-series visualization $V$\;
Synthesize $R$ and $Q$ to generate concise textual assessment $T$\;
\BlankLine
Return $V$ and $T$ to the operator frontend\;
\end{algorithm}

\section{Case Study and Results}
\label{sec:results}


\begin{table}[t]
\centering
\caption{{Batch-Level Validation of LLM-to-SQL Query Generation}}
\label{tab:batch_validation}
\scriptsize
\setlength{\tabcolsep}{3pt}
\renewcommand{\arraystretch}{1.08}
\begin{tabular}{p{0.31\columnwidth} p{0.48\columnwidth} p{0.13\columnwidth}}
\hline
\textbf{Metric} & \textbf{Definition} & \textbf{Result} \\
\hline
SQL validity & Generated query is syntactically valid PostgreSQL. & 20/20 \\
Approved-view compliance & Query uses only approved KPI views and schema fields. & 20/20 \\
Execution success & Valid query executes without database error. & 20/20 \\
Semantic correctness & Query retrieves the intended KPI, node, time window, or aggregation. & 19/20 \\
Invalid-query rejection & Unsafe or unsupported prompts are rejected by the validator. & 7/10 \\
Hallucinated field/view rate & Query references a non-existing field or unapproved view. & 30\% \\
\hline
\end{tabular}
\end{table}
In the manual baseline, the operator must identify the relevant KPI view, select the node, device, and time window, write or modify SQL filters, aggregate timestamped records, compare the returned values with operating limits, and prepare the engineering interpretation. The proposed framework automates the query formulation, retrieval, and constraint-assessment steps while preserving read-only database access and deterministic engineering checks. This comparison is functional rather than time-based; a formal operator-timing and usability study against deployed ADMS/DERMS dashboards is left for future work. The proposed framework was evaluated using telemetry collected from a hardware-in-the-loop co-simulation of a BESS operating under reactive power-based voltage control and price-driven demand response. The evaluation includes two parts. First, a batch prompt suite was used to test SQL validity, approved-view compliance, execution success, semantic correctness, invalid-query rejection, hallucinated field or view references, and rejection of database-modification requests. These aggregate LLM-to-SQL validation results are reported in Table~\ref{tab:batch_validation}. Second, four detailed case-study prompts were used to demonstrate end-to-end engineering assessment for voltage-duration analysis, voltage-vulnerable node ranking, reactive-power overshoot coincidence, and demand-response active-power tracking. Table~\ref{tab:baseline_comparison} summarizes the functional difference between the proposed workflow and a manual SQL/dashboard monitoring baseline.
These prompts were selected because each requires a different database operation, including duration reconstruction, node-level ranking, temporal coincidence checking, or run-level tracking-error aggregation. In the database, a \textit{record} refers to a single timestamped row of telemetry sampled at 1-second intervals, while a \textit{violation event} refers to a contiguous block of consecutive records during which the monitored variable remained outside its operating bound. This distinction is used consistently across all case studies below.
\begin{table}[t]
\caption{Functional comparison with a manual monitoring baseline}
\label{tab:baseline_comparison}
\centering
\footnotesize
\setlength{\tabcolsep}{3pt}
\begin{tabular}{p{0.30\columnwidth}p{0.30\columnwidth}p{0.30\columnwidth}}
\toprule
Task & Manual SQL/dashboard workflow & Proposed framework \\
\midrule
Query formulation & Operator writes SQL or selects filters & Natural-language query converted to SQL \\
Schema use & Requires view and field knowledge & Uses approved schema context \\
Data retrieval & Manual node, device, and time-window selection & Validated query retrieves required telemetry \\
Constraint check & Manual comparison with voltage, SOC, and tracking limits & Deterministic constraint layer evaluates results \\
Invalid request handling & Depends on user review or database error & Validator blocks unsupported SQL \\
Output & Manual interpretation and plotting & Text, dashboard, or plot/table assessment \\
\bottomrule
\end{tabular}
\end{table}

\subsection{Case A: Voltage Stability Duration Assessment}

The following prompt was submitted to the framework:

\par\noindent
{\itshape``For the most voltage-vulnerable node in this run, how long did it stay within and outside the acceptable voltage limits?"}

The framework generated a SQL query targeting the \texttt{v\_voltage\_violations} view. It collected the times where the voltage fell outside the 0.95--1.05~pu operating range. The framework found that the violations occured at 1,125 timestamps, identifying Node~56 as the most vulnerable location in the network, with a worst-case undervoltage of 0.9302~pu recorded at 2026-04-01 21:42:45 UTC, with a difference of 0.0198~pu below the lower bound. The duration of this violation occurred approximately for 1,000~seconds ($\approx$16.7~minutes).

The framework evaluated the result, utilizing Table~\ref{tab:constraints} as reference, to generated the following engineering assessment:

\par\noindent
{\itshape
"The most voltage-vulnerable node in this run is node\_id 56 (all three phases), which exhibits the largest deviation from the acceptable band. Based on the run metrics, it spent about 1000 seconds ($\approx$16.7 minutes) outside the acceptable voltage limits (0.95--1.05~pu). The worst observed point was an undervoltage of 0.9302~pu at 2026-04-01 21:42:45 UTC, corresponding to a 0.0198~pu deviation below the 0.95~pu limit. Condition label: \textbf{vulnerable}. Recommendation: prioritize voltage support at/near Node~56 (e.g., adjust regulator/LTC setpoints, add/retune capacitor or inverter Volt-VAR control, and check feeder loading around the 21:39--21:43~UTC interval) to reduce the duration and depth of undervoltage."}

\begin{figure}[ht]
\centering
\includegraphics[width=1\columnwidth, trim=20 10 20 10, clip]{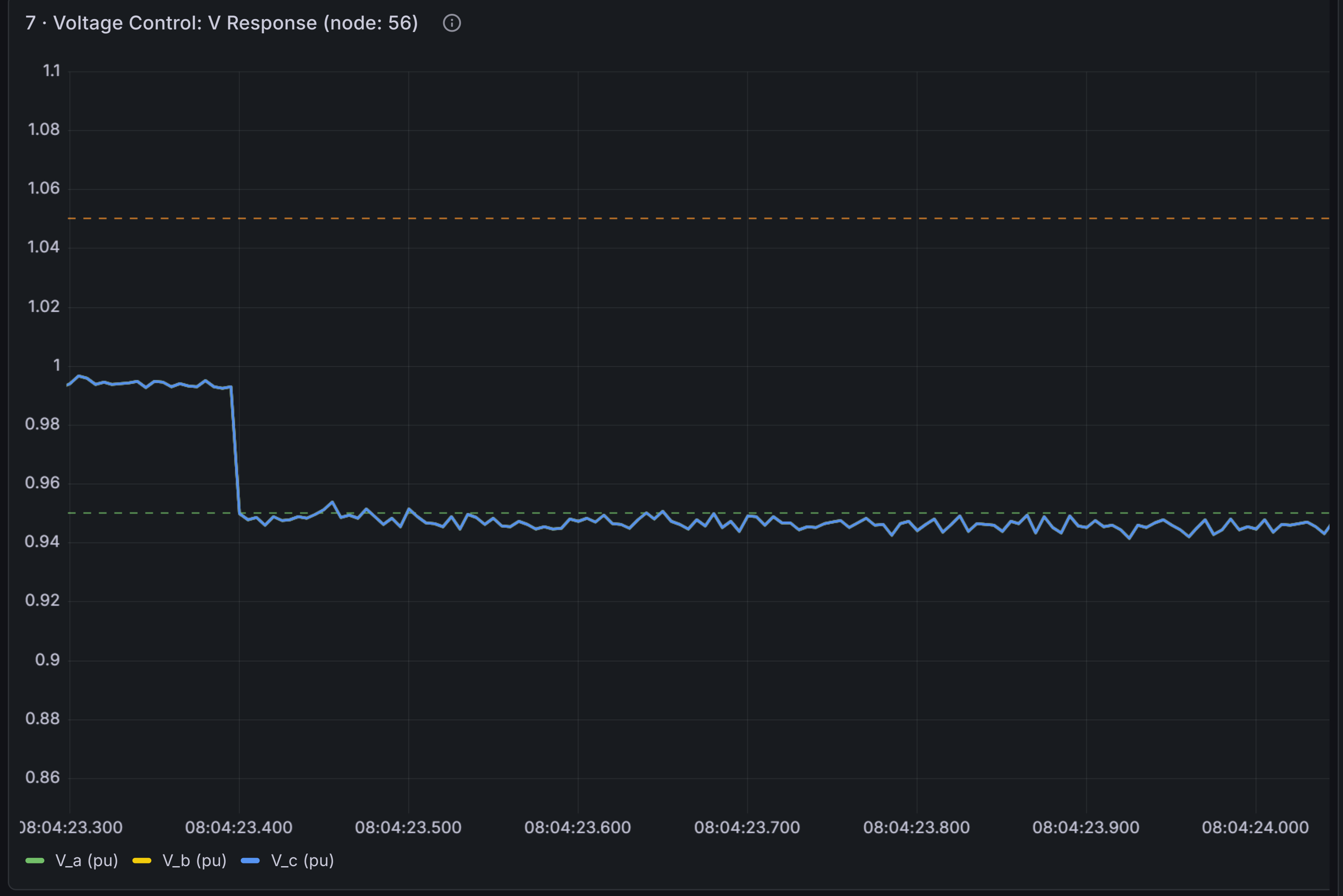}
\caption{Voltage control response at Node~56 during the evaluated run, showing sustained undervoltage excursions below the 0.95~pu lower operating bound.}
\label{fig:voltage_response}
\end{figure}

The voltage control response during this interval is shown in Fig.~\ref{fig:voltage_response}, confirming the the undervoltage condition at Node~56.

\subsection{Case B: Identification of the Most Voltage-Vulnerable Node}

The following prompt was submitted:

\par\noindent
{\itshape``Which node had the most voltage violations in this run, and what were the violation statistics?"}

The framework generated a SQL query that collects data from violation statistics across all monitored buses and ranked nodes by total violation count. The query returned one summary row per node. Node~56 was again identified as the most vulnerable node, recording 621 violation events, all of which were undervoltage (zero overvoltage events). The voltage at Node~56 ranged from 0.9302~pu to 0.9499~pu with a mean of 0.9455~pu, and a maximum single-sample deviation of 0.0198~pu. Those 621 violation events matched up with the 1,125 violation recorded from Case~A: Case~A counted every individual 1-second sample outside the voltage bound across all phases, while Case~B compiled those samples into per-node event counts, providing complementary views of the same undervoltage condition.

The framework generated the following assessment:

\par\noindent
{\itshape
``Node~56 had the most voltage violations in this run, with 621 violation events --- all undervoltage (0 overvoltage), spanning 3 phases. The voltage ranged from 0.9302~pu to 0.9499~pu with a mean of about 0.9455~pu, and the maximum deviation was 0.0198~pu. Condition label: \textbf{vulnerable}. Recommendation: investigate feeder/transformer tap settings and local reactive power support near Node~56 (e.g., capacitor/VAR control), and verify conductor/loading assumptions to bring voltages above 0.95~pu."}

\begin{figure}[ht]
\centering
\includegraphics[width=1\columnwidth, trim=20 10 20 10, clip]{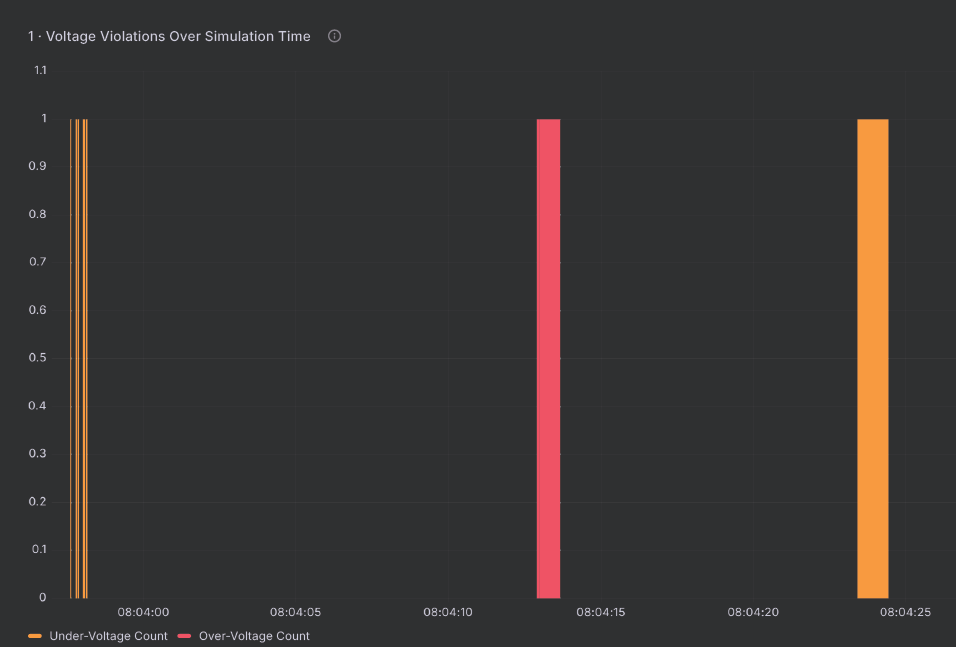}
\caption{Voltage violation event counts per node over the evaluated run, identifying Node~56 as the most frequently violated bus in the feeder.}
\label{fig:voltage_violations}
\end{figure}

The per-node violation counts across the feeder are shown in Fig.~\ref{fig:voltage_violations}, confirming Node~56 as the most violated node.

\subsection{Case C: Reactive Power Overshoot Detection}

The following prompt was submitted:

\par\noindent
{\itshape``Did reactive power overshoot coincide with voltage violations at any node during Volt-VAR control operation? Which nodes were most affected?"}

The framework generated a SQL query that combined the \texttt{v\_voltage\_control\_summary} and \texttt{v\_voltage\_violations} by matching node and time to find the time periods where \texttt{max\_q\_overshoot} exceeded zero and \texttt{voltage\_viol\_flag} was active at the same time. Only four nodes met these criteria, at each node, the number of overlapping event, the total for both voltage violations and reactive overshoot events.

Node~56 had the most overlap, with 621 events and a maximum voltage drop of 0.0198pu. Node 13 had 285 overlapping events and a maximum drop of 0.0102~pu. Node~1 recorded 186 events (maximum drop 0.0075~pu) and Node~57 recorded 33 events (maximum deviation 0.0025~pu). The constraint layer labeled the overal condition as \textit{stable} because the voltage changes were not high enough to be consider severely unstable.

The framework generated the following assessment:

\par\noindent
{\itshape
``Yes. During Volt-VAR control operation, reactive power overshoot
(\texttt{q\_violation} = TRUE) coincided with voltage violation timestamps at
4 nodes. The most affected node was Node~56 with 621 coincident events and
the largest observed voltage deviation (0.0198~pu), followed by Node~13 with
285 events (max 0.0102~pu) and Node~1 with 186 events (max 0.0075~pu).
Node~57 was least impacted with 33 coincident events (max 0.0025~pu).
Overall condition label: \textbf{stable}. Recommendation: review and retune
Volt-VAR controller gains/deadbands or reactive power limits at Nodes~56
and~13 to reduce reactive overshoot and associated voltage excursions."
}

\subsection{Case D: Demand Response Active-Power Tracking}

Following prompt was submitted:

\par\noindent
{\itshape``How well did the BESS track the active power reference during demand response operation in this run?"}

The framework generated a SQL query to check the \texttt{v\_kpi\_run\_summary} view and collect overall active-power tracking results for all Demand Response times. The query returned one summary row covering 50,000 samples from 2026-04-01 12:00:00 UTC to 2026-04-02 05:21:35 UTC. The average DR tracking error was 0.8529, much higher than the severe threshold of 0.1, indicating large and sustained deviation between the requested and active power during the whole period. Because of this, the constraint layer marked as \textit{vulnerable}.

The framework generated the following assessment:

\par\noindent
{\itshape``The BESS did not track the active power reference well during demand response in this run. Over 50,000 samples from 2026-04-01 12:00 UTC to 2026-04-02 05:21 UTC, the mean absolute DR tracking error was 0.8529 --- far above the severe threshold of 0.1. This indicates large, sustained deviation between commanded and delivered active power. Condition label: \textbf{vulnerable}. Recommendation: review and retune the active power control loop and limit handling (e.g., ramp-rate, saturation/anti-windup, and inverter power limits) to reduce tracking error, and verify telemetry scaling/units for the reference and measured power signals."}

\begin{figure}[ht]
\centering
\includegraphics[width=1\columnwidth, trim=20 10 20 10, clip]{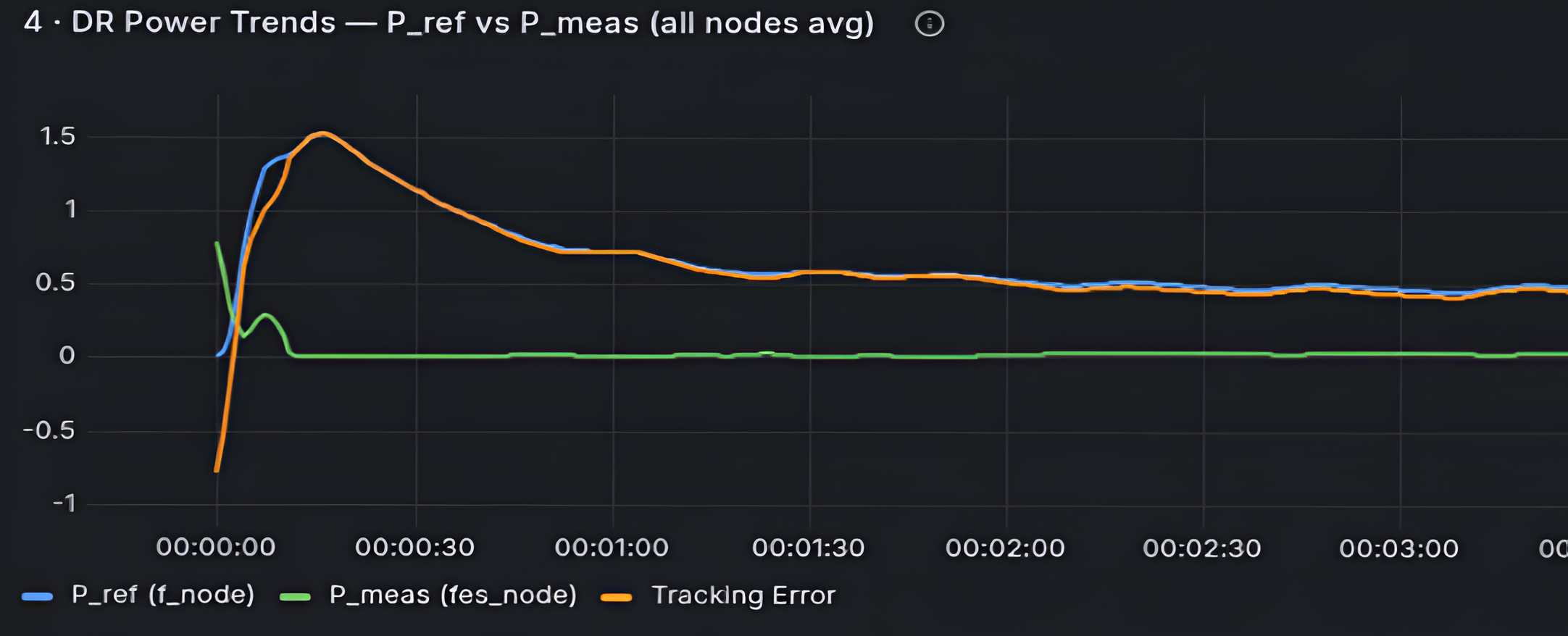}
\caption{Demand response active-power tracking during the evaluated run: reference command ($P_{\mathrm{ref}}$) vs.\ measured BESS response ($P_{\mathrm{meas}}$), showing sustained tracking deviation classified as vulnerable.}
\label{fig:dr_power}
\end{figure}

The active-power tracking performance is shown in Fig.~\ref{fig:dr_power}, which compares $P_{\mathrm{ref}}$ against $P_{\mathrm{meas}}$ across the full DR dispatch window and illustrates the magnitude of the tracking deviation identified by the framework.

\section{Conclusions}
\label{sec:conclusion}

This paper presented an AI-enabled monitoring framework integrating a large language model with a PostgreSQL telemetry database for BESS operation analysis. This pipeline allows the language model to serve as a natural-language interface, while SQL validation and deterministic constraint checks control the numerical and engineering content of the response. It translates natural-language operator questions into structured SQL queries, retrieves operational telemetry from predefined KPI views, evaluates results against engineering constraints, and generates concise engineering assessments. Case-study evaluation demonstrated successful SQL generation, accurate retrieval of operational information, and meaningful interpretation of voltage regulation and reactive power control behavior, supported by the voltage response and demand response tracking results shown in Figs.~\ref{fig:voltage_response} and~\ref{fig:dr_power}. The results indicate that LLM-assisted monitoring can provide a practical interface between telemetry databases and engineering decision-support systems for future power distribution grid applications. The functional comparison in Table~\ref{tab:baseline_comparison} shows that the proposed framework reduces the need for manual KPI-view selection, SQL writing, constraint checking, and engineering interpretation. The present comparison is not a formal user-efficiency study; future work will quantify operator time, query success, and usability against conventional dashboards, manual SQL workflows, and ADMS/DERMS monitoring interfaces. Future deployment work will also evaluate the framework on live distribution feeders so that operators can query telemetry as it is generated.
While this study demonstrates the technical feasibility of schema-constrained LLM-assisted monitoring, a formal user-efficiency comparison against conventional dashboards, manual SQL workflows, and existing ADMS/DERMS monitoring interfaces will be explored in future work. We also plan to collaborate with electric service providers to implement the system on live distribution feeders, allowing operators to query telemetry as it is generated.

\bibliographystyle{IEEEtran}
\bibliography{ref}

@ARTICLE{11168242,
  author={Mirshekali, Hamid and Reza Shadi, Mohammad and Ghanadi Ladani, Fatemehsadat and Reza Shaker, Hamid},
  journal={IEEE Access}, 
  title={A Review of Large Language Models for Energy Systems: Applications, Challenges, and Future Prospects}, 
  year={2025},
  volume={13},
  number={},
  pages={163162-163188},
  doi={10.1109/ACCESS.2025.3610994}}

@article{fose2024empowering,
  title={Empowering distribution system operators: A review of distributed energy resource forecasting techniques},
  author={Fose, Nande and Singh, Arvind R and Krishnamurthy, Senthil and Ratshitanga, Mukovhe and Moodley, Prathaban},
  journal={Heliyon},
  volume={10},
  number={15},
  year={2024},
  publisher={Elsevier}
}

@ARTICLE{8013143,
  author={Zhao, Huiying and Hong, Mingguo and Lin, Wei and Loparo, Kenneth A.},
  journal={IEEE Transactions on Smart Grid}, 
  title={Voltage and Frequency Regulation of Microgrid With Battery Energy Storage Systems}, 
  year={2019},
  volume={10},
  number={1},
  pages={414-424},
  doi={10.1109/TSG.2017.2741668}}

@INPROCEEDINGS{10318643,
  author={Peterson, Mary and O'Reilly, Olivia and Manoharan, Arun-Kaarthick and Rajendran, Sarangan and Melagoda, Adithya and Aravinthan, Visvakumar and Liu, Esther and Tamimi, Al and Yokley, Charles},
  booktitle={2023 North American Power Symposium (NAPS)}, 
  title={Economic and Reliability Impacts of Combined Solar and Battery Energy Storage as a Non-Wire Alternative}, 
  year={2023},
  volume={},
  number={},
  pages={1-6},
  doi={10.1109/NAPS58826.2023.10318643}}

@ARTICLE{10918749,
  author={Rafy, Md Fazley and Boateng, Ellis Oti and Krishnan, Vignesh Venkata Gopala and Srivastava, Anurag K.},
  journal={IEEE Transactions on Industry Applications}, 
  title={Cyber-Resilient IoT-Based Battery Energy Storage Systems in Power Distribution System}, 
  year={2025},
  volume={61},
  number={3},
  pages={4566-4577},
  doi={10.1109/TIA.2025.3549413}}

@techreport{pnnl2024connecting,
  title={Connecting Minds: AI Use Cases to Bridge Power Systems and Large Language Models for Practical Applications},
  author={Chen, Yousu and Anderson, Alexander A},
  year={2025},
  institution={Pacific Northwest National Laboratory (PNNL), Richland, WA (United States)}
}

@article{sugunaraj2025distributed,
  title={Distributed energy resource management system (DERMS) cybersecurity scenarios, trends, and potential technologies: a review},
  author={Sugunaraj, Niroop and Balaji, Shree Ram Abayankar and Chandar, Barathwaja Subash and Rajagopalan, Prashanth and Kose, Utku and Loper, David Charles and Mahfuz, Tanzim and Chakraborty, Prabuddha and Ahmad, Seerin and Kim, Taesic and others},
  journal={IEEE Communications Surveys \& Tutorials},
  volume={28},
  pages={224--277},
  year={2025},
  publisher={IEEE}
}

@article{spudic2022integration,
  title={Integration of utility distributed energy resource management system and aggregators for evolving distribution system operators},
  author={Strezoski, Luka and Padullaparti, Harsha and Ding, Fei and Baggu, Murali},
  journal={Journal of Modern Power Systems and Clean Energy},
  volume={10},
  number={2},
  pages={277--285},
  year={2022},
  publisher={SGEPRI}
}

@article{wu2022distribution,
  title={Distribution control centers in the US and Europe: Commonalities, differences, and lessons},
  author={Vadari, Subramanian and D{\v{z}}afi{\'c}, Izudin and Koch, Dan'l and Murphy, Ryan and Hayes, Daniel and Donlagic, Tarik},
  journal={Journal of Modern Power Systems and Clean Energy},
  volume={10},
  number={2},
  pages={259--268},
  year={2022},
  publisher={SGEPRI}
}

@article{cobilean2023review,
  title={A review of visualization methods for cyber-physical security: Smart grid case study},
  author={Cobilean, Victor and Mavikumbure, Harindra S and Mcbride, Brady J and Vaagensmith, Bjorn and Singh, Vivek Kumar and Li, Ruixuan and Rieger, Craig and Manic, Milos},
  journal={IEEE Access},
  volume={11},
  pages={59788--59803},
  year={2023},
  publisher={IEEE}
}

@ARTICLE{11457738,
  author={Rafy, MF and Sharma, P. and Patari, N. and Srivastava, A. K. and Sharma, A.},
  journal={IEEE Transactions on Industry Applications}, 
  title={Edge-Driven Distributed Control for Power Distribution: A Real-Time Hardware-in-The-Loop Testbed for Industrial Automation and Applications}, 
  year={2026},
  volume={},
  number={},
  pages={1-12},
  doi={10.1109/TIA.2026.3678576}}

@inproceedings{sarwar2022characterization,
  title={Characterization and mitigation of fault induced delayed voltage recovery with dynamic voltage support by hybrid pv plants},
  author={Sarwar, Muhammad and Matavalam, Amarsagar Reddy Ramapuram and Ajjarapu, Venkataramana},
  booktitle={2022 North American Power Symposium (NAPS)},
  pages={1--6},
  year={2022},
  organization={IEEE}
}

@inproceedings{wei2023chance,
  title={A chance-constrained optimal design of Volt/VAR control rules for distributed energy resources},
  author={Wei, Jinlei and Gupta, Sarthak and Aliprantis, Dionysios C and Kekatos, Vassilis},
  booktitle={2023 North American Power Symposium (NAPS)},
  pages={1--6},
  year={2023},
  organization={IEEE}
}

@inproceedings{kunkolienkar2024visualizing,
  title={Visualizing Volt-Var Distributions in Large-Scale Electric Grid Models},
  author={Kunkolienkar, Sanjana and Cook, Jordan and Overbye, Thomas J},
  booktitle={2024 56th North American Power Symposium (NAPS)},
  pages={1--6},
  year={2024},
  organization={IEEE}
}

@inproceedings{jin2024chatgrid,
  title={ChatGrid: Power grid visualization empowered by a large language model},
  author={Jin, Sichen and Abhyankar, Shrirang},
  booktitle={2024 IEEE Workshop on Energy Data Visualization (EnergyVis)},
  pages={12--17},
  year={2024},
  organization={IEEE}
}

@article{bonadia2023potential,
  title={On the potential of ChatGPT to generate distribution systems for load flow studies using OpenDSS},
  author={Bonadia, Rodrigo S and Trindade, Fernanda CL and Freitas, Walmir and Venkatesh, Bala},
  journal={IEEE Transactions on Power Systems},
  volume={38},
  number={6},
  pages={5965--5968},
  year={2023},
  publisher={IEEE}
}

@article{jena2025llm,
  title={LLM-Based Adaptive Distribution Voltage Regulation Under Frequent Topology Changes: An In-Context MPC Framework},
  author={Jena, Amit and Ding, Fei and Wang, Jiyu and Yao, Yiyun and Xie, Le},
  journal={IEEE Transactions on Smart Grid},
  year={2025},
  publisher={IEEE}
}

@article{hong2025next,
  title={Next-generation database interfaces: A survey of llm-based text-to-sql},
  author={Hong, Zijin and Yuan, Zheng and Zhang, Qinggang and Chen, Hao and Dong, Junnan and Huang, Feiran and Huang, Xiao},
  journal={IEEE Transactions on Knowledge and Data Engineering},
  year={2025},
  publisher={IEEE}
}

@techreport{walker2023battery,
  title={Battery Energy Storage System Evaluation Method},
  author={Walker, Andy and Desai, Jal},
  year={2023},
  institution={National Renewable Energy Laboratory (NREL), Golden, CO (United States)}
}

@article{ruan2024applying,
  title={Applying large language models to power systems: Potential security threats},
  author={Ruan, Jiaqi and Liang, Gaoqi and Zhao, Huan and Liu, Guolong and Sun, Xianzhuo and Qiu, Jing and Xu, Zhao and Wen, Fushuan and Dong, Zhao Yang},
  journal={IEEE transactions on smart grid},
  volume={15},
  number={3},
  pages={3333--3336},
  year={2024},
  publisher={IEEE}
}

\end{document}